\documentclass[letterpaper, 10 pt, conference]{ieeeconf}  %
\pdfoutput=1
\usepackage[T1]{fontenc}

\IEEEoverridecommandlockouts                              %
\def\abovestrut#1{\rule[0in]{0in}{#1}\ignorespaces}
\def\belowstrut#1{\rule[-#1]{0in}{#1}\ignorespaces}
\usepackage{textcase}
\usepackage{subcaption}
\usepackage{hyperref}
\usepackage{graphicx} %
\usepackage{booktabs} 
\usepackage{amssymb}
\usepackage{times}
\usepackage{epsfig}
\usepackage{graphicx}
\usepackage{color}
\usepackage{url}
\usepackage{hyperref}
\usepackage{bbm}
\usepackage{multicol}

\providecommand{\hide}[1]{}

\usepackage{amsmath,amsfonts}
\usepackage{amsopn,amssymb}

\usepackage{amsthm}

\theoremstyle{plain}
\newtheorem{thm}{Theorem}%

\newtheorem{cor}[thm]{Corollary}

\theoremstyle{definition}

\theoremstyle{remark}

\usepackage{bm} %
\usepackage{algorithmicx}
\usepackage{algorithm}%
\PassOptionsToPackage{noend}{algpseudocode}
\usepackage{algpseudocode}%
\algnewcommand{\LineComment}[1]{\Statex \(\triangleright\) #1}

\newcommand{\vct}[1]{\boldsymbol{#1}} %
\newcommand{\mat}[1]{\boldsymbol{#1}} %
\newcommand{\T}{^{\textrm T}} %

\newcommand{\ProbOpr}[1]{\mathbb{#1}}
\newcommand{\expect}[2]{%
\ifthenelse{\equal{#2}{}}{\ProbOpr{E}_{#1}}
{\ifthenelse{\equal{#1}{}}{\ProbOpr{E}\left[#2\right]}{\ProbOpr{E}_{#1}\left[#2\right]}}} %
\newcommand{\var}[2]{%
\ifthenelse{\equal{#2}{}}{\ProbOpr{VAR}_{#1}}
{\ifthenelse{\equal{#1}{}}{\ProbOpr{VAR}\left[#2\right]}{\ProbOpr{VAR}_{#1}\left[#2\right]}}} %
\DeclareMathOperator{\argmax}{arg\,max}

\newcommand{\vx}{{\vct{x}}}

\newcommand{\vy}{\vct{y}}

\newcommand{\vk}{\vct{k}}

\newcommand{\vxi}{\vct{\xi}}

\newcommand{\mI}{\mat{I}}
\newcommand{\mK}{\mat{K}}

   \newcommand{\cd}{\mathcal{D}}

\algnewcommand{\algorithmicgoto}{\textbf{go to}}%
\algnewcommand{\Goto}[1]{\algorithmicgoto~\ref{#1}}%

\usepackage{xspace}
\newcommand{\rej}{{\sc{rejection}}\xspace}
\newcommand{\lse}{{\sc gp-lse}\xspace}
\newcommand{\diverse}{{\sc{diverse}}\xspace}
\newcommand{\adapt}{{\sc{adaptive}}\xspace}
\newcommand{\gk}{{\sc{diverse-gk}}\xspace}
\newcommand{\lk}{{\sc{diverse-lk}}\xspace}

\newcount\Comments  %
\usepackage{array,multirow,graphicx}
\usepackage{color}
\usepackage{wrapfig}
\definecolor{darkgreen}{rgb}{0,0.5,0}
\definecolor{purple}{rgb}{1,0,1}
\newcommand{\kibitz}[2]{\ifnum\Comments=1\textcolor{#1}{#2}\fi}
\newcommand{\zw}[1]{\kibitz{purple}      {[ZW: #1]}}

\newcommand{\gp}{{\sc gp}}
\newcommand{\GP}{{\rm GP}}  %
\newcommand{\stripstream}{{\sc strips}tream}
\newcommand{\tamp}{{\sc tamp}}
\newcommand{\dpp}{{\sc dpp}}

\title{\LARGE \bf
Active model learning and diverse action sampling \\ for 
 task and motion planning 
}
\author{Zi Wang \quad Caelan Reed Garrett \quad Leslie Pack Kaelbling \quad Tom\'as Lozano-P\'erez%
\thanks{* MIT CSAIL. 
        {\tt \{ziw,caelan,lpk,tlp\}@csail.mit.edu}. 
        We gratefully acknowledge support from NSF grants 1420316, 1523767 and 1723381, from AFOSR grant FA9550-17-1-0165, from Honda Research and Draper Laboratory.  Any opinions, findings, and conclusions or recommendations expressed in this material are those of the authors and do not necessarily reflect the views of our sponsors.\newline\vskip 0.02in \noindent\emph{Proceedings of the 2018 IEEE/RSJ International Conference on Intelligent Robots and Systems (IROS), Madrid, Spain.}}%
}

\begin{document}

\maketitle
\thispagestyle{empty}
\pagestyle{empty}

\begin{abstract}
The objective of this work is to augment the basic abilities of a robot by learning to use new sensorimotor primitives to enable the solution of complex long-horizon problems. Solving long-horizon problems in complex domains requires flexible generative planning that can combine primitive abilities in novel combinations to solve problems as they arise in the world.  In order to plan to combine primitive actions, we must have models of the preconditions and effects of those actions: under what circumstances will executing this primitive achieve some particular effect in the world?

We use, and develop novel improvements on, state-of-the-art methods for active learning and sampling.  We use Gaussian process methods for learning the conditions of operator effectiveness from small numbers of expensive training examples collected by experimentation on a robot. We develop adaptive sampling methods for generating diverse elements of continuous sets (such as robot configurations and object poses) during planning for solving a new task, so that planning is as efficient as possible. We demonstrate these methods in an integrated system, combining newly learned models with an efficient continuous-space robot task and motion planner to learn to solve long horizon problems more efficiently than was previously possible.

\end{abstract}

\section{Introduction} 
\label{sec:intro}

For a robot to be effective in a domain that combines novel
sensorimotor primitives, such as pouring or stirring, with
long-horizon, high-level task objectives, such as cooking a meal or 
making a cup of coffee, it 
is necessary to acquire models of these primitives to use in planning
robot motions and manipulations.  These models characterize (a)
conditions under which the primitive is likely to succeed and (b) the
effects of the primitive on the state of the world.  

\begin{figure}
\centering
\includegraphics[width=\columnwidth]{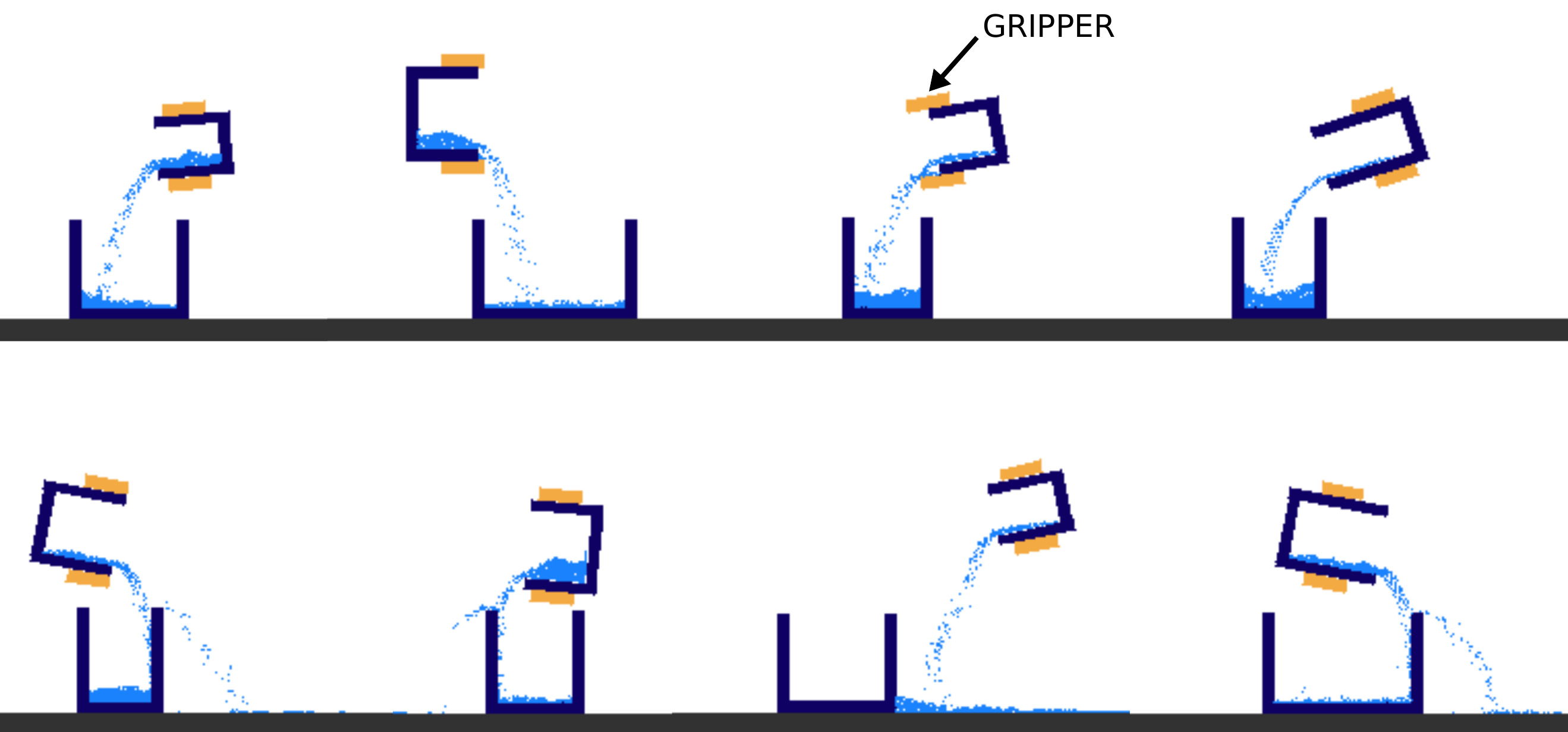}
\vspace{-1em}
\caption{Several examples of executing a pouring primitive with
  different settings, including control parameters, cup sizes, and
  relative placements.}
  \vspace{-1.5em}
\label{fig:pouring}
\end{figure}

Figure~\ref{fig:pouring} illustrates several instances of a
parameterized motor primitive for pouring in a simple two-dimensional
domain.  The primitive action has control parameters $\theta$ that
govern the rate at which the cup is tipped and target velocity of the
poured material.  In addition, several properties of the situation in
which the pouring occurs are very relevant for its success: robot
configuration $c_R$, pouring cup pose and size $p_A, s_A$, and target
cup pose and size $p_B, s_B$.  To model the effects of the action we
need to specify $c_R'$ and $p_A'$, the resulting robot configuration
and pose of the pouring cup $A$.  Only for some settings of the parameters
$(c_R, p_A, s_A, p_B, s_B, \theta, c'_R, p'_A) \in \chi$ is the action
feasible: one key objective of our work is to efficiently learn a
representation of the feasible region $\chi$.

For learning this model, each training example requires running the
primitive, which is expensive on real robot hardware and even in
high-fidelity simulation.  To minimize the amount of training data
required, we actively select each setting in which the primitive is
executed, with the goal of obtaining as much information as possible
about how to use the primitive.  This results in a dramatic reduction
in required examples over our preliminary
work~\cite{kaelbling2017learning} on this problem.

Given a model of a primitive, embodied in $\chi$, we utilize existing
sample-based algorithms for {\em task and motion planning} (\tamp{})
to find plans.  To use the model within the planner, it is necessary
to select candidate instances of the primitives for expansion during
the search for a plan.  The objective here is not to gain information,
but to select primitive instances that are likely to be successful.
It is not enough to select a single instance, however, because there
may be other circumstances that make a particular instance infeasible
within a planning context: for example, although the most reliable way
to grasp a particular object might be from the top, the robot might
encounter the object situated in a cupboard in a way that makes the
top grasp infeasible.  Thus, our action-sampling mechanism must select
instances that are both likely to succeed and are {\em diverse} from
one another, so that the planner has coverage of the space of possible
actions.

One difficulty in sampling $\chi$ is that it inhabits a
lower-dimensional submanifold of the space it is defined in, because
some relations among robot configurations and object poses, for
example, are functional.  The \stripstream{}
planner~\cite{GarrettRSS17, garrett2017strips} introduced a strategy
for sampling from such {\em dimensionality-reducing} constraints by
constructing {\em conditional samplers} that, given values of some
variables, generate values of the other variables that satisfy the
constraint.  Our goal in this paper is to learn and use conditional
samplers within the \stripstream{} planner.

Our technical strategy for addressing the problems of (a) learning
success constraints and (b) generating diverse samples is based on an
explicit representation of uncertainty about an underlying {\em scoring}
function that measures the quality or likelihood of success of a
parameter vector, and uses Gaussian process ({\sc gp}) techniques to
sample for information-gathering during learning and for success
probability and diversity during planning.  We begin by describing
some basic background, discuss related work, describe our methods in
technical detail, and then present experimental results of learning
and planning with several motor primitives in a
two-dimensional dynamic simulator.

\section{Problem formulation and background}

We will focus on the formal problem of learning and using a
conditional sampler of the form  $\phi(\theta \mid \alpha)$,
where $\alpha\in R^{d_{\alpha}}$ is a vector of contextual parameters
and $\theta\in B$ is a vector of parameters that are to be generated,
conditioned on $\alpha$.  
We assume in the following that the domain of $\theta$ is a
hyper-rectangular space $B=[0,1]^{d_\theta}\subset R^{d_{\theta}}$,
but generalization to other topologies is possible.
The conditional sampler 
generates samples of $\theta$ such that $(\theta, \alpha) \in \chi$
where $\chi \subset R^{d_{\alpha}+d_{\theta}}$ characterizes the set of
world states and parameters for which the skill is feasible.
 We assume that $\chi$ can be expressed in the 
form of an inequality constraint $\chi(\theta, \alpha) = (g(\theta,
\alpha) > 0)$, where $g(\cdot)$ is a scoring function with arguments
$\theta$ and $\alpha$. We denote the {\em super level-set} of the scoring
function given $\alpha$ by $A_{\alpha}=\{\theta \in B\mid g(\theta, \alpha) >
0\}$. For example, the scoring function $g(\cdot)$ for pouring might
be the proportion of poured liquid that actually ends up in the target
cup, minus some target proportion.
We assume the availability of values of such a score function during
training rather than just binary labels of success or failure.
In the following, we give basic background on two important components
of our method:  Gaussian processes and \stripstream{}.

\label{ssec:gp}
Gaussian processes (\gp{}s) are distributions over functions, and
popular priors for Bayesian non-parametric regression. 
In a \gp{}, any finite set of function values has a multivariate
Gaussian distribution. In this paper, we use the Gaussian process
$\GP(0,k)$ which has mean zero and covariance (kernel) function
$k(\vx,\vx')$.   
\hide{
Two frequently used examples of covariance kernel functions are the squared exponential and Mat\'ern kernels.
Let $r=(\vx-\vx')^\top(\vx-\vx')$. Then the squared exponential kernel is 
$$k(\vx,\vx') = \sigma_f^2 \mathrm e^{-\frac{1}{2\ell^2} r},$$ 
with hyper-parameters $(\sigma_f,\ell)$. The Mat\'ern kernel is given by 
$$k(\vx,\vx') = \sigma_m^2\frac{2^{1-\xi}}{\Gamma(\xi)} (\frac{\sqrt{2\xi r}}{h})^{\xi}B_\xi(\frac{\sqrt{2\xi r}}{h}),$$ 
where $\Gamma$ is the gamma function, $B_\xi$ is a modified Bessel
function, and we have the hyper-parameters $\sigma_m$, $h$, and the
roughness parameter $\xi$. 
}
Let $f$ be a true underlying function sampled from $\GP(0,k)$. 
Given a set of observations $\cd=\{(\vx_t,y_t)\}_{t=1}^{|\cd|}$, 
where $y_t$ is an evaluation of $f$ at $\vx_t$ 
corrupted by i.i.d additive Gaussian noise $\mathcal N(0,\zeta^2)$,
we obtain a posterior \gp{}, with 
mean  
$\mu(\vx) = \vk^{\cd}(\vx)\T(\mK^\cd+\zeta^2\mI)^{-1}\vy^\cd$ 
and covariance  
$\sigma^2(\vx, \vx') = k(\vx,\vx') -
\vk^\cd(\vx)\T(\mK^\cd+\zeta^2\mI)^{-1} \vk_t(\vx')$ 
where the kernel matrix $\mK^\cd
=\left[k(\vx_i,\vx_j)\right]_{\vx_i,\vx_j\in \cd}$ and $\vk^\cd(\vx) =
[k(\vx_i,\vx)]_{\vx_i\in \cd}$~\cite{rasmussen2006gaussian}. 
With slight abuse of notation, we denote the posterior variance by
$\sigma^2(\vx) = \sigma^2(\vx,\vx)$, and the posterior \gp{} by $\GP(\mu,
\sigma)$.

\stripstream{}~\cite{garrett2017strips} is a framework for incorporating blackbox sampling procedures in a planning language. 
It extends the STRIPS planning language~\cite{Fikes71} by adding {\em streams}, declarative specifications of conditional generators. 
Streams have previously been used to model off-the-shelf motion planners, collision checkers, inverse kinematic solvers.
In this work, we learn new conditional generators, such as samplers for pouring, and incorporate them using streams.

\section{Related Work}

Our work draws ideas from model learning, probabilistic modeling of
functions, and task and motion planning (\tamp{}).

There is a large amount of work on learning individual motor
primitives such pushing~\cite{kroemer2016meta, hermans2013learning},
scooping~\cite{schenck2017learning}, and
pouring~\cite{pan2016robot,tamosiunaite2011learning,brandi2014generalizing,yamaguchi2016differential,schenck2017visual}.
We focus on the task of learning models of these primitives suitable
for multi-step planning.  We extend a particular formulation of
planning model learning~\cite{kaelbling2017learning}, where
constraint-based pre-image models are learned for parameterized action
primitives, by giving a probabilistic characterization of the
pre-image and using these models during planning.

Other approaches exist to learning models of the preconditions and
effects of sensorimotor skills suitable for planning.
One~\cite{konidaris2018skills} constructs a completely symbolic model
of skills that enable purely symbolic task planning.  Our method, on
the other hand, learns hybrid models, involving continuous
parameters. Another~\cite{kroemer2016learning} learns image
classifiers for pre-conditions but does not support general-purpose
planning.

We use \gp-based level set estimation~\cite{bryan2006active, gotovos2013active,
  rasmussen2006gaussian, bogunovic2016truncated} to model the feasible
regions (super level set of the scoring function) of action
parameters.  We use the {\em straddle} algorithm~\cite{bryan2006active}
to actively sample from the function threshold, in order to estimate
the super level set that satisfy the constraint with high probability.
Our methods can be extended to other function approximators that gives
uncertainty estimates, such as Bayesian neural
networks and their variants~\cite{gal2016dropout,lakshminarayanan2016simple}.

Determinantal point processes ({\dpp}s)~\cite{kulesza2012determinantal} are typically used for diversity-aware sampling.
However, both sampling from a continuous \dpp{}~\cite{hafiz2013approximate} and learning the kernel of a \dpp{}~\cite{affandi2014learning} are challenging. 

Several approaches to \tamp{} utilize generators to enumerate infinite
sequences of
values~\cite{kaelbling2011hierarchical,srivastava2014combined,GarrettRSS17}.
Our learned samplers can be incorporated in any of these approaches.
Additionally, some recent papers have investigated learning effective
samplers within the context of \tamp{}.  Chitnis et
al.~\cite{chitnis2016guided} frame learning plan parameters as a
reinforcement learning problem and learn a randomized policy that
samples from a discrete set of robot base and object poses.  Kim et
al.~\cite{kimICRA17} proposed a method for selecting from a discrete
set of samples by scoring new samples based on their correlation with
previously attempted samples.  In subsequent work, they instead train
a Generative Adversarial Network to directly produce a distribution of
satisfactory samples~\cite{kimAAAI2018}.

\section{Active sampling for learning and planning}
\label{sec:method}

Our objective in the learning phase is to efficiently gather data
to characterize the conditional super-level-sets $A_\alpha$ with high
confidence.  We use a \gp{} on the score function $g$ to select
informative queries using a level-set estimation approach.
Our objective in the planning phase is to select
a diverse set of samples $\{\theta_i\}$ for which it is likely that
$\theta \in A_\alpha$.  We do this in two steps:  first, we use a
novel risk-aware sampler to generate $\theta$ values 
that satisfy the constraint with high probability; second, we
integrate this sampler with \stripstream{}, where we generate
samples from this set that represent its diversity, in order to 
expose the full variety of choices to the planner.

\subsection{Actively learning the constraint with a GP}

Our goal is to be able to sample from the super level set
$A_{\alpha}=\{\theta \in B\mid g(\theta, \alpha) > 0\}$ for any given
context $\alpha$, which requires learning the decision 
boundary $g(\theta, \alpha) = 0$.   During training, we select
$\alpha$ values from a
distribution reflecting naturally occurring contexts in the underlying
domain.  
Note that learning the level-set is
a different objective from learning all of the function values
well, and so it must be handled differently from typical \gp{}-based
active learning.

For each $\alpha$ value in the training set, we apply the {\em
  straddle} algorithm~\cite{bryan2006active} to actively select
samples of $\theta$ for evaluation by running the motor primitive.
After each new evaluation of $g(\theta, \alpha)$ is obtained, the
data-set $\mathcal D$ is augmented with pair
$\langle (\theta, \alpha), g(\theta, \alpha) \rangle$, and used to
update the \gp{}.  The straddle algorithm selects $\theta$ that
maximizes the acquisition function
$\psi(\theta; \alpha, \mu, \sigma) = -|\mu (\theta,\alpha)|+1.96\sigma
(\theta,\alpha)$. It has a high value for values of $\theta$ that
are near the boundary for the given $\alpha$ or for which the score
function is highly uncertain.  The parameter $1.96$ is selected such
that if $\psi(\theta)$ is negative, $\theta$ has less than 5 percent
chance of being in the level set.  In practice, this heuristic has
been observed to deliver state-of-the-art learning performance for
level set estimation~\cite{bogunovic2016truncated,gotovos2013active}.
After each new evaluation, we retrain the Gaussian process by
maximizing its marginal data-likelihood with respect to its
hyper-parameters.  Alg.~\ref{alg:bo} specifies the algorithm;
GP-predict$(\cdot)$ computes the posterior mean and variance 
as explained in Sec.~\ref{ssec:gp}.

\begin{algorithm}[H]
  \begin{footnotesize}
  \caption{Active Bayesian Level Set Estimation}\label{alg:bo}
  \begin{algorithmic}[1]
    \State Given initial data set $\mathcal D$, context $\alpha$,
    desired number of samples $T$
      \For{$t = 1\rightarrow T$}
      \State $\mu, \sigma$ $\gets$ GP-predict($\cd$)
      \State $\theta\gets \argmax_{\theta}{\psi(\theta; \alpha, \mu, \sigma)}$ %
      \State $y\gets g(\theta, \alpha)$ %
      \State $\cd \gets \cd \cup \{\left( (\theta,\alpha),y\right) \}$
      \EndFor
     \State \Return{$\mathcal D$}
  \end{algorithmic}
  \end{footnotesize}
\end{algorithm}

\subsection{Risk-aware adaptive sampling for constraint satisfaction} 
\label{ssec:adaptive}
Now we can use this Bayesian estimate of the scoring function
$g$ to select action instances for planning.  Given a
new context $\alpha$, which need not have occured in the training
set---the \gp{} will provide generalization over contexts---we would
like to sample a sequence of $\theta\in B$ such that with high
probability, $g(\theta, \alpha)\geq 0$. In order to guarantee this, we
adopt a concentration bound and a union bound on the predictive scores
of the samples. Notice that by construction of the \gp{}, the predictive
scores are Gaussian random variables.  The following is a
direct corollary of Lemma 3.2 of~\cite{wang2016est}.  

\begin{cor}\label{lem:pbound}
Let $g(\theta) \sim \GP(\mu, \sigma)$, $\delta\in(0,1)$ and
set $\beta^*_i = (2\log(\pi_i / 2\delta))^\frac12$, where
$\sum_{i=1}^T \pi_i^{-1} \leq 1$, $\pi_i > 0$. If  $\mu(\theta_i) >
\beta^*_i\,\sigma(\theta_i), \forall  i=1,\cdots, T$, then 
$ \Pr [ g(\theta_i) > 0, \forall i ] \geq 1-\delta$.
\end{cor}  

Define the high-probability super-level-set of $\theta$ given context
$\alpha$ as 
$\hat A_\alpha=\{\theta \mid \mu(\theta, \alpha) > \beta^* \,
\sigma(\theta, \alpha) \}$ where $\beta^*$ is picked according to
Corollary~\ref{lem:pbound}.  If we draw $T$ samples from $\hat A_\alpha$,
then with probability at least $1-\delta$, all of the samples will
satisfy the constraint $g(\theta, \alpha) > 0$.  

In practice, however, for any given $\alpha$, using the definition of
$\beta^*$ from Corollary~\ref{lem:pbound}, the set $\hat A_\alpha$ may be
empty.    In that case, we can relax our criterion to include the set
of $\theta$ values whose score is within 5\% of the $\theta$ value
that is currently estimated to have the highest likelihood of
satisfying the constraint: 
$\beta = \Phi^{-1}(0.95 \Phi(\max_{\theta\in B} \mu(\theta, \alpha) / \sigma(\theta,
\alpha)))$ where $\Phi$ is
the cumulative density function of a normal distribution.

\hide{
Let
$\beta_* = \max_{\theta\in B} \mu(\theta, \alpha) / \sigma(\theta,
\alpha)$.
Because
$\mu(\theta, \alpha) \leq \beta_*\,\sigma(\theta, \alpha), \forall
\theta\in B$, we have
$\mu(\theta, \alpha)< \beta\,\sigma(\theta, \alpha), \forall \theta\in
B$ if $\beta > \beta_*$.
As a result, $\Phi(\mu(\theta, \alpha) /
  \sigma(\theta, \alpha))$ is the probability that
  $g(\theta,\alpha)>0$. So $\Phi(\beta_*)$ is the highest probability
  for a certain $\theta\in B$ such that $g(\theta,\alpha)>0$. This
  means $\beta_*$ can serve an indicator of how confident the
  algorithm is. 

If  $\Phi(\beta_*)$ is too small, the sampler should
  really return an empty answer, although our current implementation
  does not support this.

Instead, to ensure the existence of the region, we relax the criterion by 
using an operator $\lambda(\cdot)$ such that
$\lambda(\beta_*) < \beta_*$, and use $\beta = \lambda(\beta_*)$ to
construct $\hat A_\alpha$.  
For example, one can use
$\lambda(\beta_*) = \Phi^{-1}(0.95 \Phi(\beta_*)) $ where $\Phi$ is
the cumulative density function of a normal distribution. The
interpretation is that the $\theta$ to be sampled has at least $95\%$
of the probability that the best $\theta$ has of satisfying the
constraint.
}

\begin{figure}
\centering
\vskip 0.1in
\includegraphics[width=.95\columnwidth]{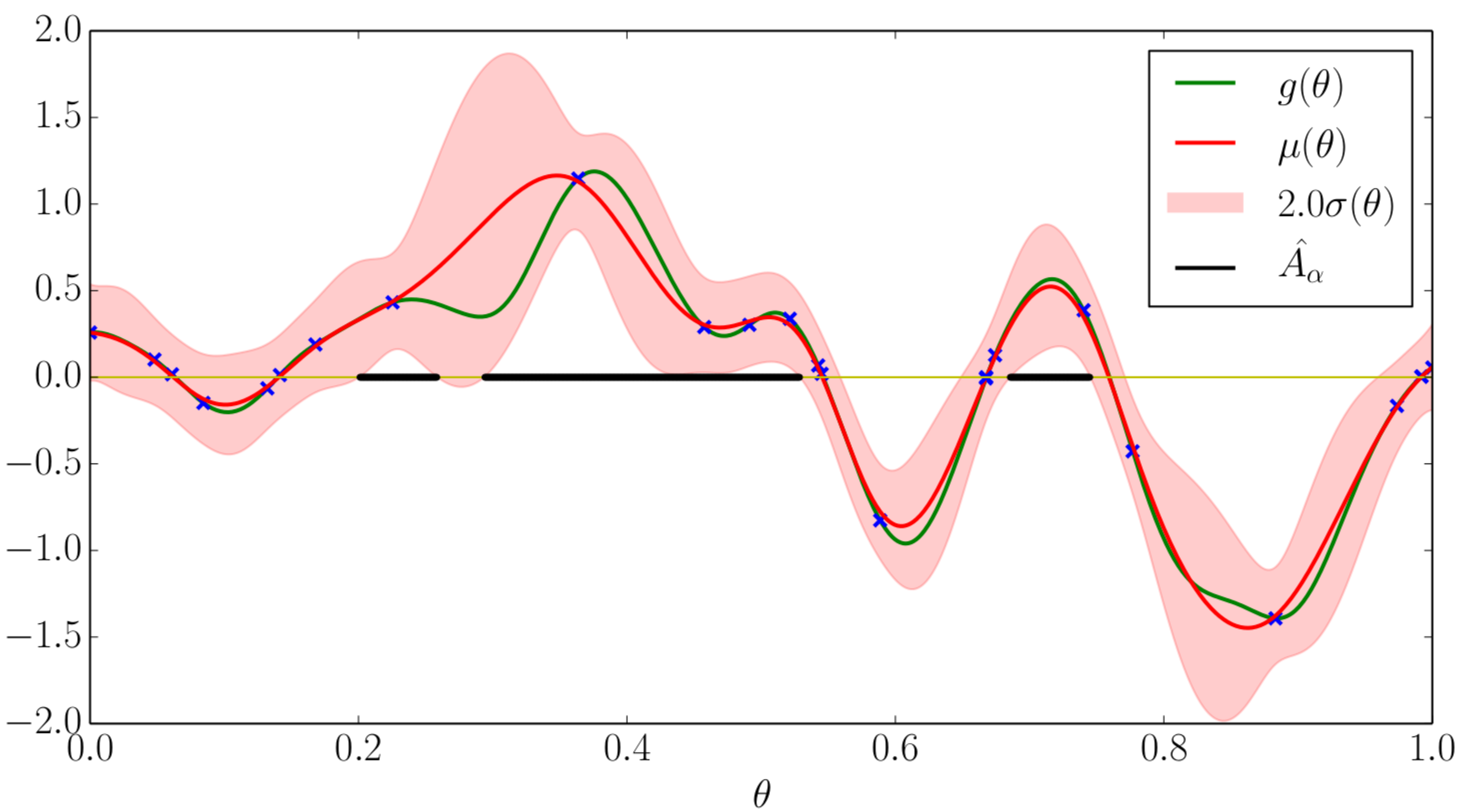}
\caption{High-probability super-level-set in black.}
\label{fig:hpsls}
\vskip -0.2in
\end{figure}

Figure~\ref{fig:hpsls} illustrates the computation of $\hat A_\alpha$.
The green line is 
the true hidden $g(\theta)$; the blue $\times$ symbols are the training
data, gathered using the straddle algorithm in $[0,1]$; the red
line is the posterior mean function $\mu(\theta)$; the pink
regions show the two-standard-deviation bounds on $g(\theta)$ based
on $\sigma(\theta)$; and the black line segments are the
high-probability super-level-set $\hat A_\alpha$ for $\beta = 2.0$.
We can see that sampling has concentrated near the boundary, that 
$\hat A_\alpha$ is a subset of the true super-level-set, and that as
$\sigma$ decreases through experience, $\hat A_\alpha$ will approach
the true super-level set.

To sample from $\hat A_{\alpha}$, one simple strategy is to do
rejection sampling with a proposal distribution that is uniform on the
search bounding-box $B$.  However, in many
cases, the feasible region of a constraint is typically much smaller
than $B$, which means that uniform sampling will have a very low
chance of drawing samples within $\hat A_\alpha$, and so rejection
sampling will be very inefficient.
We address this problem using a novel adaptive sampler, which 
draws new samples from the neighborhood of
the samples that are already known to be feasible with high probability
and then re-weights these new samples using importance 
weights. 

The algorithm {\sc AdaptiveSampler} takes as input the posterior \gp{}
parameters $\mu$ and $\sigma$ and context vector $\alpha$, and yields
a stream of samples.
It begins by computing $\beta$ and sets $\Theta_{\it init}$ to contain
the $\theta$ that is most likely to satisfy the constraint.
It then maintains a buffer $\Theta$ of at least $m/2$ samples, and
yields the first one each time it is required to do so;  it
technically never actually returns, but generates 
a sample each time it is called. 
The main work is done by {\sc SampleBuffer}, which constructs a
mixture of truncated Gaussian distributions ({\sc tgmm}), specified by
mixture weights $p$, means $\Theta$, circular variance with parameter
$v$, and bounds $B$. 
Parameter $v$  indicates how far from known good $\theta$ values it is reasonable
to search;  it is increased if a large portion
of the samples from the {\sc tgmm} are accepted and decreased otherwise.
The algorithm iterates until it has constructed a set of at least $m$
samples from $\hat A_\alpha$.  It samples $n$ elements from the {\sc
  tgmm} and retains those that are in $\hat A_\alpha$ as $\Theta_a$.
Then, it computes ``importance weights'' $p_a$ that are inversely
related to the probability of drawing each $\theta_a \in \Theta_a$
from the current {\sc tgmm}.  This will tend to spread the mass of the
sampling distribution away from the current samples, but still
concentrated in the target region.  A set of $n$ uniform
samples is drawn and filtered, again to maintain the chance of
dispersing to good regions that are far from the initialization.  The
$p$ values associated with the old $\Theta$ as well as the newly
sampled ones are concatenated and then normalized into a distribution,
the new samples added to $\Theta$, and the loop continues.  When at
least $m$ samples have been obtained, $m$
elements are sampled from $\Theta$ according to distribution $p$,
without replacement.

\begin{algorithm} %
  \begin{footnotesize}
  \caption{Super Level Set Adaptive Sampling}\label{alg:adaptive}
  \begin{algorithmic}[1]
  \Function{AdaptiveSampler}{$\mu, \sigma, \alpha$}
 \State $\beta \gets \Phi^{-1}(0.95 \Phi(\max_{\theta\in B}
 \mu(\theta, \alpha) / \sigma(\theta, \alpha)))$
     \State $\Theta_{init}\gets \{\argmax_{\theta\in
       B}\frac{\mu(\theta)}{\sigma(\theta)} \}$; $\Theta \gets \emptyset$
     \While{{\bf True}}
          \If{$|\Theta| < m/2$}
          \State $\Theta \gets \textsc{SampleBuffer}(\mu, \sigma, \alpha, \beta, \Theta_{init}, n, m)$
          \EndIf
          \State $\theta \gets \Theta[0]$
          \State \textbf{yield} $\theta$
          \State $\Theta \gets \Theta\setminus \{\theta\}$
          \EndWhile
  \EndFunction
  \Function{SampleBuffer}{$\mu, \sigma, \alpha, \beta, \Theta_{init}$}
  \State $v\gets [1]_{d=1}^{d_{\theta}}$;  $\Theta\gets \Theta_{init}$; $p \gets [1]_{i=1}^{|\Theta|}$
      \While{True}
       \State $\Theta' \gets$ SampleTGMM$(n ; p, \Theta,v, B)$
      \State $\Theta_{a} \gets\{\theta \in\Theta' \mid \mu(\theta) > \beta \sigma(\theta)\}$
      \State $p_{a} \gets 1/p_{\text{TGMM}}(\Theta_{a}; p, \Theta, v, B)$
      \State $v \gets v / 2  \;{\bf if}\; |\Theta_{a}| < |\Theta'|/2 \;{\bf else}\; v \times 2$
      
      \State $\Theta'' \gets$ SampleUniform$(n; B)$
      \State $\Theta_{r} \gets$ $\{\theta \in\Theta'' \mid \mu(\theta) > \beta \sigma(\theta)\}$

      \State $p_{r} \gets [Vol(B)]_{i=1}^{|\Theta_{r}|}$ %
      \State $p\gets$ Normalize$([p, p_{r}, p_{a}])$ 
      \State $\Theta \gets $ $[\Theta, \Theta_{r}, \Theta_{a}]$
      \If{$|\Theta| > m$}
      \State \Return Sample$(m; \Theta, p)$
      \EndIf
      \EndWhile
   \EndFunction
  \end{algorithmic}
  \end{footnotesize}
\end{algorithm}
  \vskip -0.05in
It is easy to see that as $n$ goes to infinity, by sampling from the
discrete set according to the re-weighted probability, we are
essentially sampling uniformly at random from $\hat A_{\alpha}$.
This is because $\forall \theta \in
\Theta, p(\theta) \propto
\frac{1}{p_{sample}(\theta)}p_{sample}(\theta) = 1$. For
uniform sampling, $p_{sample}(\theta) = \frac{1}{Vol(B)}$, where $Vol(B)$ 
is the volume of $B$; and for
sampling from the truncated mixture of Gaussians, $p_{sample}(\theta)$
is the probability density of $\theta$. In practice, $n$ is finite,
but this method is much more efficient than rejection sampling.

\subsection{Diversity-aware sampling for planning}
\label{ssec:diverse}
Now that we have a sampler that can generate approximately
uniformly random samples within the region of values that satisfy the
constraints with high probability, we can use it inside a planning
algorithm for continuous action spaces.  Such planners perform
backtracking search, potentially needing to consider multiple
different parameterized instances of a particular action before
finding one that will work well in the overall context of the planning
problem.  The efficiency of this process depends on the order in which
samples of action instances are generated.
Intuitively, when previous samples of this action for this context
have failed to contribute to a successful plan, it would be wise to
try new samples that, while still having high probability of
satisfying the constraint, are as different from those that were
previously tried as possible.
We need, therefore, to consider diversity when generating samples; but
the precise characterization of useful diversity depends on the domain
in which the method is operating.   We address this problem by
adapting a kernel that is used in the sampling process, based on
experience in previous planning problems.

Diversity-aware sampling has been studied extensively with
determinantal point processes ({\sc
  dpp}s)~\cite{kulesza2012determinantal}. We begin with similar ideas
and adapt them to the planning domain, quantifying diversity of a 
set of samples $S$ using the determinant of a Gram matrix:
$D(S) = \log \det(\Xi^S\zeta^{-2} + \mI)$, where
$\Xi^{S}_{ij} = \xi(\theta_i, \theta_j), \forall \theta_i,\theta_j\in
S$, $\xi$ is a covariance function, and $\zeta$ is a free parameter
(we use $\zeta=0.1$). In {\sc dpp}s, the quantity $D(S)$ can be interpreted
as the volume spanned by the feature space of the kernel
$\xi(\theta_i, \theta_j)\zeta^{-2} + \textbf{1}_{\theta_i\equiv
  \theta_j}$ assuming that $\theta_i=\theta_j \iff
i=j$. Alternatively, one can interpret the quantity $D(S)$ as the
information gain of a \gp{} when the function values on $S$ are
observed~\cite{srinivas2009gaussian}. This \gp{} has kernel $\xi$ and
observation noise $\mathcal N(0,\zeta^2)$. 
Because of the submodularity and
monotonicity of $D(\cdot)$, we can maximize $D(S)$ greedily with the
promise that
$D([\theta_i]_{i=1}^N)\geq (1-\frac1e)\max_{|S|\leq N}D(S)$
$\forall N=1,2,\cdots$, where
$\theta_i = \argmax_{\theta} D(\theta \cup
\{\theta_j\}_{j=1}^{i-1})$. In fact, maximizing $D(\theta \cup S)$ is
equivalent to maximizing
\[
\eta_S(\theta) = \xi(\theta,\theta) - \vxi^{S}(\theta)\T(\Xi^{S}+\zeta^2\mI)^{-1}\vxi^{S}(\theta)
\]
 which is exactly the same as the posterior variance for a \gp{}.

\begin{algorithm}[H]
  \begin{footnotesize}
  \caption{Super Level Set Diverse Sampling}\label{alg:diverse}
  \begin{algorithmic}[1]
  \Function{DiverseSampler}{$\mu, \sigma, \alpha, \eta$}
  \State $\beta \gets \lambda(\max_{\theta\in
    B}\frac{\mu(\theta)}{\sigma(\theta)})$; $\Theta \gets \emptyset$
  \State $\theta \gets \argmax_{\theta\in
    B}\frac{\mu(\theta)}{\sigma(\theta)}$; $S\gets \emptyset$
      \While{planner requires samples}
      \State \textbf{yield} $\theta$, S
      \If{$|\Theta| < m/2$}
      \State $\Theta \gets \textsc{SampleBuffer}(\mu, \sigma, \alpha, \beta, \Theta_{init})$
      \EndIf
      \State $S\gets S\cup\{\theta\}$
      \Comment $S$ contains samples before $\theta$
      \State $\theta \gets \argmax_{\theta\in\Theta} \eta_{S} (\theta)$
      \State $\Theta \gets \Theta\setminus \{\theta\}$
      \EndWhile
  \EndFunction
  \end{algorithmic}
  \end{footnotesize}
\end{algorithm}
\vskip -0.1in
The {\sc DiverseSampler} procedure is very similar in structure to the
{\sc AdaptiveSampler} procedure, but rather than
selecting an arbitrary element of $\Theta$, the buffer of good
samples, to return, we track the set $S$ of samples that have already
been returned and select the element of $\Theta$ that is most diverse
from $S$ as the sample to yield on each iteration.  In addition, we
yield $S$ to enable kernel learning as described in
Alg~\ref{alg:learn}, to yield a kernel $\eta$.

It is typical to learn the kernel parameters of a \gp{} or \dpp{} given
supervised training examples of function values or diverse sets, 
but those are not available in our setting;  we can only observe which samples are
accepted by the planner and which are not.
We derive our notion of similarity by assuming that all samples that
are rejected by the planner are similar. %
Under this assumption, we develop an online learning
approach that adapts the kernel parameters to learn a good diversity
metric for a sequence of planning tasks.

We use the squared exponential kernel of the form
$\xi(\theta, \gamma; l) = \exp(-\sum_d r^2_d)$, where
$r_d = |l_d(\theta_d - \gamma_d)|$ is the rescaled ``distance''
between $\theta$ and $\gamma$ on the $d$-th feature and $l$ is the
inverse lengthscale. Let $\theta$ be the sample that failed and the
set of samples sampled before $\theta$ be $S$. We define the
importance of the $d$-th feature as
\[
\tau_{S}^{\theta}(d) = \xi(\theta_d, \theta_d; l_d) - \vxi^{S}(\theta_d; l_d)\T(\Xi^{S}+\zeta^2\mI)^{-1}\vxi^S(\theta_d; l_d)\;\;,
\]
which is the conditional variance if we ignore the distance
contribution of all other features except the $d$-th; that is,
$\forall k\neq d, l_k=0$. Note that we keep $\Xi_i+\zeta^2\mI$ the
same for all the features so that the inverse only needs to be
computed once.

The diverse sampling procedure 
is analogous to the weighted majority
algorithm~\cite{foster1999regret} in that each feature $d$
is seen as an expert that contributes to the conditional
variance term, which measures how diverse $\theta$ is with respect to
$S$. The contribution of feature $d$ is measured by
$\tau^\theta_S(d)$. If $\theta$ was rejected by the planner,
we decrease the inverse lengthscale $l_d$ of feature
$d=\argmax_{d\in[d_{\theta}]}\tau_S^{\theta}(d)$ to be
$(1-\epsilon)l_d$, 
because feature $d$ contributed the most to the
decision that $\theta$ was most different
from $S$. 

\begin{algorithm}[H]
  \begin{footnotesize}
  \caption{Task-level Kernel Learning}\label{alg:learn}
  \begin{algorithmic}[1]
  \For{task in T}
    \State $\alpha \gets $ current context
    \State $\mu, \sigma$ $\gets$ GP-predict($\alpha$);  $S\gets \emptyset$
      \While{plan not found}
      \If{$|S| > 0$}
      \State $d\gets \argmax_{d\in[d_{\theta}]}\tau_S^{\theta}(d)$
      \State $l_d \gets  (1-\epsilon)l_d$
      \EndIf
      \State $\theta,S  \gets \textsc{DiverseSampler}(\mu, \sigma,
      \alpha, \xi(\cdot, \cdot; l))$
      \State Check if a plan exist using $\theta$
      \EndWhile
  \EndFor
  \end{algorithmic}
  \end{footnotesize}
  
\end{algorithm}
\vspace{-1em}
Alg.~\ref{alg:learn} depicts a scenario in which the kernel is
updated during interactions with a planner; it is simplified in that it
uses a single sampler, but in our experimental applications there are
many instances of action samplers in play during a single execution of
the planner. 
Given a sequence of tasks presented to the planner, we can continue to
apply this kernel update, molding our diversity measure to the demands
of the distribution of tasks in the domain.
This simple strategy for kernel learning may lead to a significant
reduction in planning time, as we demonstrate in the next section. 

\hide{
If the generator is asked to generate one more time, it means some other constraints are not satisfied (\zw{is it true?}), what we can do is to initialize the sampler for this particular planning problem instance with the generator we learned, and update the score of previously generated parameters to be something smaller (e.g. a threshold below 0). Doing this allows us to customize the generator for each planning problem instance. (\zw{potential problem: this is eliminating part of the level sets one at a time.. What if all level sets are removed? What to generate in that case? How to consider the task-level contextual information?} 
}

\section{Experiments}

We show the effectiveness and efficiency of each component of our
method independently, and then demonstrate their collective
performance in the context of planning for long-horizon tasks in a
high-dimensional continuous domain.

\begin{figure*}
\vskip 0.1in
\centering
\includegraphics[width=1.9\columnwidth]{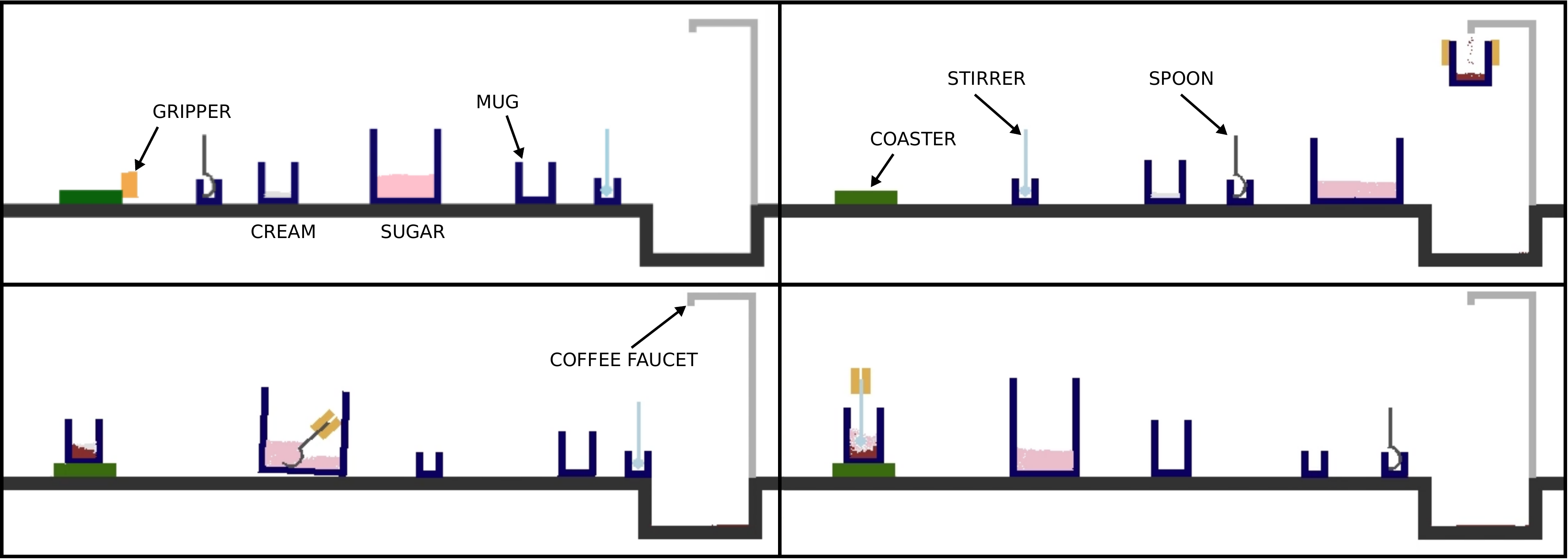}
\caption{Four arrangements of objects in 2D kitchen, including: green
  coaster, coffee faucet, yellow robot grippers, sugar scoop, stirrer,
coffee mug, small cup with cream, larger container with pink sugar.}
\label{fig:kitchen2d}
\vskip -0.2in
\end{figure*}

To test our algorithms, we implemented a simulated 2D kitchen based on
the physics engine Box2D~\cite{box2d}. Fig.~\ref{fig:kitchen2d}
shows several scenes indicating the variability of arrangements of
objects in the domain.
We use bi-directional RRT~\cite{KuffnerLaValle} 
to implement motion planning. %
The parameterized primitive
motor actions are: moving the robot (a simple ``free-flying'' hand),
picking up an object, placing an object, pushing an object, filling
a cup from a faucet, pouring a material out of a cup, 
scooping material into a spoon, and dumping material from a spoon. 
The gripper has 3 degrees of freedom (2D position and rotation). The
material to be poured or scooped is simulated as small circular
particles. 

We learn models and samplers for three of these action primitives:
pouring (4 context parameters, 4 predicted parameters, scooping (2
context parameters, 7 predicted parameters), and pushing (2 context
parameters, 6 predicted parameters).  The actions are represented by a
trajectory of way points for the gripper, relative to the object it is
interacting with.  For pouring, we use the scoring function
$g_{\it pour}(x) = \exp(2*(x*10 - 9.5)) - 1$, where $x$ is the
proportion of the liquid particles that are poured into the target
cup. The constraint $g_{\it pour}(x)>0$ means at least $95\%$ of the
particles are poured correctly to the target cup.  The context of
pouring includes the sizes of the cups, with widths ranging from $3$
to $8$ (units in Box2D), and heights ranging from $3$ to $5$.  For
scooping, we use the proportion of the capacity of the scoop that is
filled with liquid particles, and the scoring function is
$g_{\it scoop}(x) = x-0.5$, where $x$ is the proportion of the spoon
filled with particles.  We fix the size of the spoon and learn the
action parameters for different cup sizes, with width ranging from $5$
to $10$ and height ranging from $4$ to $8$.  For pushing, the scoring
function is $g_{\it push}(x) = 2-\|x-x_{\it goal}\|$ where $x$ is the
position of the pushed object after the pushing action and
$x_{\it goal}$ is the goal position; here the goal position is the
context. The pushing action learned in Sec.~\ref{ssec:exp_active} has
the same setting as~\cite{kaelbling2017learning}, viewing the
gripper/object with a bird-eye view.  We will make the code for the
simulation and learning methods public at {\small
  \url{https://github.com/zi-w/Kitchen2D}}.

\subsection{Active learning for conditional samplers}
\label{ssec:exp_active}
We demonstrate the performance of using a \gp{} with the straddle
algorithm (\lse) to estimate the level set 
of the constraints on parameters for pushing, pouring and scooping. 
For comparison, we also implemented a simple
method~~\cite{kaelbling2017learning}, which uses a neural network to
map $(\theta, \alpha)$ pairs to predict the probability of success
using a logistic output.   Given a partially trained network and 
a context $\alpha$, the $\theta^* = \argmax_\theta {\rm NN}(\alpha,
\theta)$ which has the highest probability of success with $\alpha$ is
chosen for execution.  Its success or failure is observed, and then
the network is retrained with this added data point.  This method is
called ${\rm NN}_c$ in the results.  In addition, we implemented a
regression-based variation that predicts $g(\theta, \alpha)$ with a
linear output layer, but given an $\alpha$ value still chooses the
maximizing $\theta$.  This method is called ${\rm NN}_r$.  We also
compare to random sampling of $\theta$ values, without any training.

\begin{figure*}[t]
\vskip 0.1in
\centering
\includegraphics[width=1\textwidth]{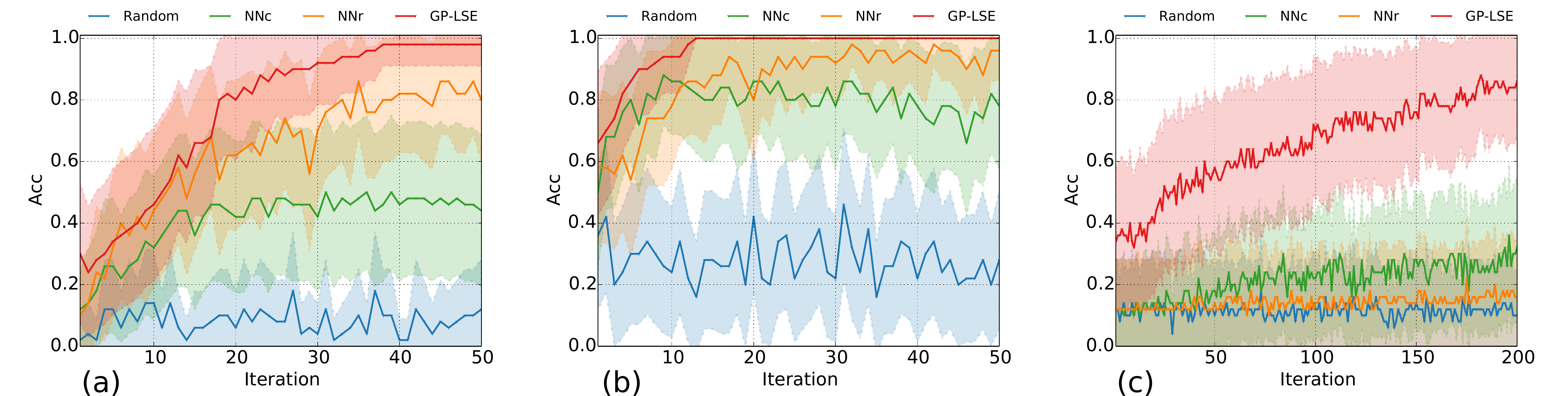}
\caption{Mean accuracy (with 1/2 stdev on mean shaded)  
  of the first action recommended by each method. 
}
\label{fig:active}
\vskip -0.1in
\end{figure*}

\lse is able to learn much more efficiently than the other methods.
Fig.~\ref{fig:active} shows the accuracy of the first action parameter
vector $\theta$ (value 1 if the action with parameters $\theta$ is
actually successful and 0 otherwise) recommended by each of these
methods as a function of the number of actively gathered training
examples.  {\sc gp-lse} recommends its first $\theta$ by maximizing
the probability that $g(\theta, \alpha) > 0$. The neural-network
methods recommend their first $\theta$ by maximizing the output value,
while {\sc random} always selects uniformly randomly from the domain
of $\theta$.  In every case, the \gp{}-based method achieves perfect
or high accuracy well before the others, demonstrating the
effectiveness of uncertainty-driven active sampling methods.

\subsection{Adaptive sampling and diverse sampling}
Given a probabilistic estimate of a desirable set of $\theta$ values,
obtained by a method such as \lse, the next step is to sample values
from that set to use in planning.  We compare simple rejection
sampling using a uniform proposal distribution (\rej), the basic
adaptive sampler from Sec.~\ref{ssec:adaptive}, and the
diversity-aware sampler from Sec.~\ref{ssec:diverse} with a fixed
kernel:  the results are shown in Table.~\ref{tb:sampling}.

\begin{figure}
\centering
\includegraphics[width=1.0\columnwidth]{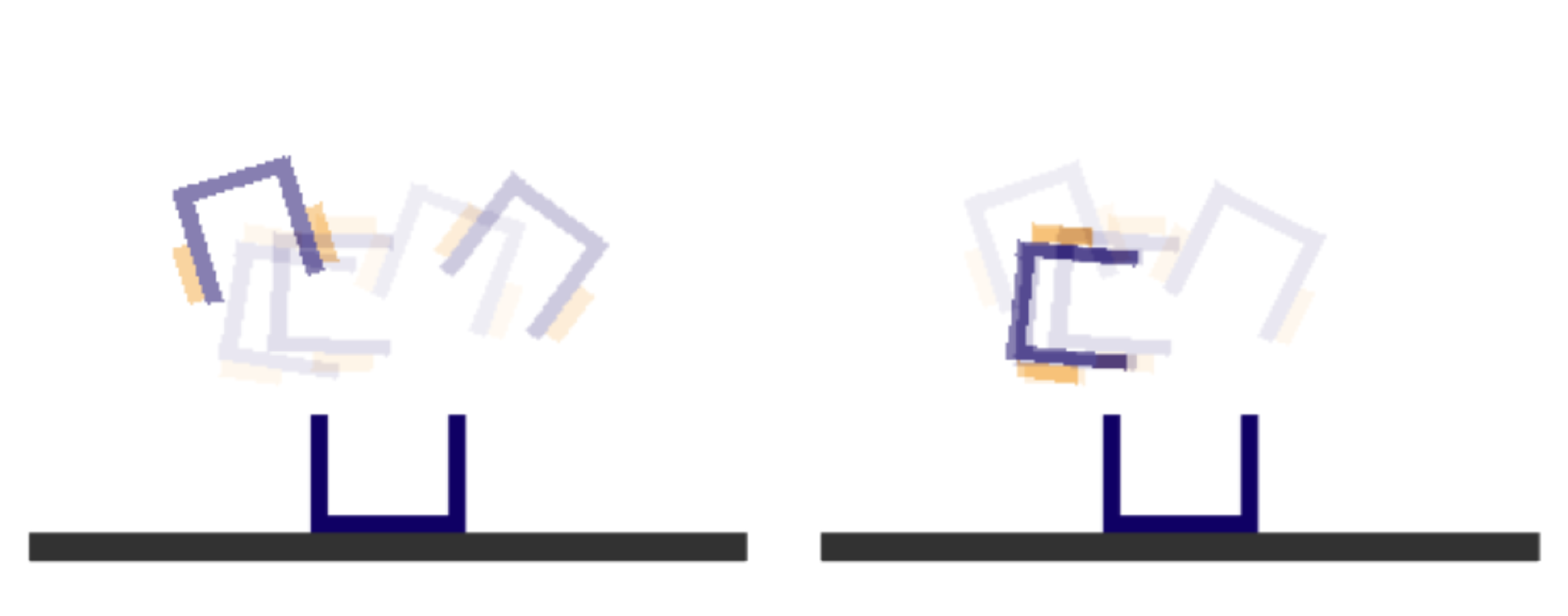}
\caption{Comparing the first 5 samples generated by \diverse
  (left) and \adapt (right) on one of the
  experiments for pouring. The more transparent the pose, the later it
  gets sampled.} 
\label{fig:gp-lse-d}
\vskip -0.1in
\end{figure}

\begin{table}
\caption{
Effectiveness of adaptive and diverse sampling.
} 
\label{tb:sampling}
\begin{center}
\begin{footnotesize}
\begin{tabular}{llccc}%
\hline
\abovestrut{0.15in}\belowstrut{0.10in}

&   & \rej & \adapt & \diverse  \\
\hline
\abovestrut{0.10in}
\parbox[t]{0mm}{\multirow{4}{*}{\rotatebox[origin=c]{90}{Pour}}}& FP (\%) & $6.45\pm 8.06$* & {\color{red}$4.04\pm 6.57$} & $5.12\pm 6.94$ \\
& $T_{50}$ (s) & $3.10\pm 1.70$* &  $0.49\pm 0.10$ & $0.53\pm 0.09$ \\
& $N_5$ & $5.51\pm 1.18$* &$5.30\pm 0.92$ & $5.44\pm 0.67$ \\
& Diversity & $17.01\pm 2.90$* & $16.24\pm 3.49$ &  {\color{red}$18.80\pm 3.38$} \\
\hline
\abovestrut{0.10in}
\parbox[t]{0mm}{\multirow{4}{*}{\rotatebox[origin=c]{90}{Scoop}}}& FP (\%) & $0.00^{\dagger}$ & {\color{red}$2.64\pm 6.24$} & $3.52\pm 6.53$ \\
& $T_{50}$ (s) & $9.89\pm 0.88^{\dagger}$ &  $0.74\pm 0.10$ & $0.81\pm 0.11$ \\
& $N_5$ & $5.00^{\dagger}$ & $5.00\pm 0.00$ & $5.10\pm 0.41$ \\
& Diversity & $21.1^{\dagger}$ & $20.89\pm 1.19$ & {\color{red}$21.90\pm 1.04$} \\
\hline
\abovestrut{0.10in}
\parbox[t]{0mm}{\multirow{4}{*}{\rotatebox[origin=c]{90}{Push}}}& FP (\%) & $68.63\pm 46.27^\ddagger$ & {\color{red}$21.36\pm 34.18$} & $38.56\pm 37.60$ \\
& $T_{50}$ (s) & $7.50\pm 3.98^\ddagger$ &  $3.58\pm 0.99$ & $3.49\pm 0.81$ \\
& $N_5$ & $5.00\pm 0.00^\ddagger$ & $5.56\pm 1.51^\triangle$ & $6.44\pm 2.11^\clubsuit$  \\
& Diversity &  $23.06\pm 0.02^\ddagger$ & $10.74\pm 4.92^\triangle$ & $13.89\pm 5.39^\clubsuit$ \\
\hline
\end{tabular}
\end{footnotesize}
\end{center}
*1 out of 50 experiments failed (to generate 50 samples within $10$ seconds);
${}^\dagger$49 out of 50 failed;
${}^\ddagger$34 out of 50 failed;
5 out of 16 experiments failed (to generate 5 positive samples within
$100$ samples); 
${}^\triangle$7 out of 50 failed;
${}^\clubsuit$11 out of 50 failed.
\vskip -0.2in
\end{table}

We report the false positive rate (proportion of samples that do not
satisfy the true constraint) on $50$ samples (FP), the time to sample
these $50$ samples ($T_{50}$), the total number of samples required to
find $5$ positive samples ($N_5$), and the diversity of those $5$
samples.  We limit cpu time for gathering $50$ samples to $10$ seconds
(running with Python 2.7.13 and Ubuntu 14.04 on Intel(R) Xeon(R) CPU
E5-2680 v3 @ 2.50GHz with 64GB memory.) If no sample is returned
within $10$ seconds, we do not include that experiment in the reported
results except the sampling time.  Hence the reported sampling time
may be a lower bound on the actual sampling time.  The diversity term
is measured by $D(S) = \log \det(\Xi^S\zeta^{-2} + \mI)$ using a
squared exponential kernel with inverse lengthscale
$l=[1, 1, \cdots, 1]$ and $\zeta=0.1$. We run the sampling algorithm
for an additional 50 iterations (a maximum of 100 samples in total)
until we have 5 positive examples and use these samples to report
$D(S)$. We also report the total number of samples needed to achieve
$5$ positive ones ($N_p$). If the method is not able to get
$5$ positive samples within $100$ samples, we report failure and do not
include them in the diversity metric or the $N_p$ metric.

\diverse uses slightly more samples than
\adapt to achieve 5 positive ones, and its false positive rate
is slightly higher than \adapt, but the diversity of the samples
is notably higher.  The FP rate of diverse can be
decreased by increasing the confidence bound on the level set.
We illustrate the ending poses of the 5 pouring actions generated by
adaptive sampling with \diverse and \adapt in
Fig.~\ref{fig:gp-lse-d}illustrating that \diverse is able to generate
more diverse action parameters, which may facilitate planning.

\hide{
\begin{table}[h]
\caption{
Comparison of methods for sampling from constraint for scooping primitive.
} 
\label{tb:scoop}
\vskip -0.2in
\begin{center}
\begin{footnotesize}
\begin{tabular}{lccc}%
\hline
\abovestrut{0.20in}\belowstrut{0.10in}

Method & \rej* & \adapt & \diverse  \\
\hline
\abovestrut{0.20in}
False Pos (\%) & $0.0$ & $2.64\pm 6.24$ & $3.52\pm 6.53$ \\
Sample time (s) & $9.89\pm 0.88$ &  $0.74\pm 0.10$ & $0.81\pm 0.11$ \\
$N_p$ & $5.00$ & $5.0\pm 0.0$ & $5.1\pm 0.41$ \\
Diversity & $21.1$ & $20.89\pm 1.19$ & {\color{red}$21.90\pm 1.04$} \\
\hline
\end{tabular}
\end{footnotesize}
\end{center}
*49 out of 50 experiments failed to generate 50 samples within $10$
seconds. 
\end{table}

\begin{table}[h]
\caption{
Comparison of methods for sampling from constraint for pushing primitive.
} 
\label{tb:push}
\vskip -0.2in
\begin{center}
\begin{footnotesize}
\begin{tabular}{lccc}%
\hline
\abovestrut{0.20in}\belowstrut{0.10in}

Method & \rej* & \adapt** & \diverse***  \\
\hline
\abovestrut{0.20in}
False Pos (\%) & $68.63\pm 46.27$ & $21.36\pm 34.18$ & $38.56\pm 37.60$ \\
Sample time (s) & $7.50\pm 3.98$ &  $3.58\pm 0.99$ & $3.49\pm 0.81$ \\
$N_p$ & $5.0\pm 0.0$ & $5.56\pm 1.51$ & $6.44\pm 2.11$  \\
Diversity &  $23.06\pm 0.02$ & $10.74\pm 4.92$ & $13.89\pm 5.39$ \\
\hline
\end{tabular}
\end{footnotesize}
\end{center}
*34 out of 50 experiments failed to generate 50 samples within $10$ seconds;
5 out of 16 experiments failed to generate 5 positive samples within $100$ samples.
**7 out of 50 experiments failed to generate 5 positive samples within $100$ samples.
***11 out of 50 experiments failed to generate 5 positive samples within $100$ samples.
\end{table}
}
\hide{
\begin{table}[h]
\caption{The false positive rate of $50$ samples,  total sampling time of the $50$
samples, number of samples to
  achieve $5$ positive ones, and the diversity rate of the 5 positive
  samples on the pushing task (repeated over 50 such samplers for each
  method). This table is using $\Phi^{-1}(0.998 \Phi(\beta_*))$. \zw{Which table to use for pushing? 
  Or remove the table for pushing??}} 
\label{tb:pour}
\vskip -0.2in
\begin{center}
\begin{footnotesize}
\begin{tabular}{lccc}%
\hline
\abovestrut{0.20in}\belowstrut{0.10in}

Method & \rej* & \adapt** & \diverse***  \\
\hline
\abovestrut{0.20in}
False Pos (\%) & $59.80\pm 48.83$ & $21.64\pm 34.18$ &  $23.16\pm 34.79$ \\
Sample time (s) & $8.30\pm 3.61$ &  $4.15\pm 1.28 $ & $3.87\pm 1.18$ \\
$N_p$ & $5.0\pm 0.0$ & $5.47\pm 1.45$ & $5.51\pm 1.17$  \\
Diversity &  $23.07\pm 0.01$ & $10.76\pm 4.91$ & $12.15\pm 5.44$ \\
\hline
\end{tabular}
\end{footnotesize}
\end{center}
*40 out of 50 experiments failed to generate 50 samples within $10$ seconds with \rej.
*6 out of 10 experiments failed to generate 5 positive samples within $100$ samples with \rej.
**7 out of 50 experiments failed to generate 5 positive samples within $100$ samples with \adapt.
*** The same 7 out of 50 experiments failed to generate 5 positive samples within $100$ samples with \diverse.
\end{table}
}

\subsection{Learning kernels for diverse sampling in planning}

\label{ssec:exp_plan}

\begin{table}
\caption{Effect of distance metric learning on sampling.}
\label{tb:timing}
\vskip -0.3in
\begin{center}
\begin{footnotesize}
\begin{tabular}{lccc}%
\hline
\abovestrut{0.15in}\belowstrut{0.10in}
 Task I& Runtime (ms) & 0.2s SR  (\%) & 0.02s SR (\%) \\
\hline
\abovestrut{0.10in}
\adapt    & $8.16\pm 12.16$  & $100.0\pm 0.0$& $87.1\pm 0.8$ \\
\gk & $9.63\pm 9.69$        & $100.0\pm 0.0$ & $82.2\pm 1.2 $\\
\lk  & {\color{red}   $5.87\pm4.63$} & $100.0\pm 0.0$ &  {\color{red}$99.9\pm 0.1$}       \\
\hline
\abovestrut{0.15in}\belowstrut{0.10in}
Task II & Runtime (s) & 60s SR  (\%) & 6s SR (\%) \\
\hline
\abovestrut{0.10in}
\adapt  & $3.22\pm 6.51$  & $91.0\pm 2.7$& $82.4\pm 5.6$ \\
\gk & $2.06\pm 1.76$        &  {\color{red} $95.0\pm 1.8 $} & $93.6\pm 2.2 $\\
\lk  & {\color{red}$1.71\pm 1.23$} & {\color{red} $95.0\pm 1.8$} &  {\color{red}$94.0\pm 1.5$}       \\
\hline
\abovestrut{0.15in}\belowstrut{0.10in}
Task III & Runtime (s) & 60s SR  (\%) & 6s SR (\%) \\
\hline
\abovestrut{0.10in}
\adapt   & $5.79\pm 11.04$  & $51.4\pm 3.3$& $40.9\pm 4.1$ \\
\gk & $3.90\pm 5.02$        &  $56.3\pm 2.0 $ & $46.3\pm 2.0 $\\
\lk  & $4.30\pm 6.89$ & {\color{red} $59.1\pm 2.6$} &  {\color{red}$49.1\pm 2.6$}       \\
\hline
\end{tabular}
\end{footnotesize}
\end{center}
\vskip -0.3in
\end{table}

In the final set of experiments, we explore the effectiveness of the
diverse sampling algorithm with task-level kernel learning 
We compare \adapt, \gk with a fixed kernel, and diverse sampling with
learned kernel (\lk), in every case using a high-probability
super-level-set estimated by a~\gp. In \lk, we use $\epsilon=0.3$.
\begin{figure*}
\vskip 0.1in
\centering
\includegraphics[width=1\textwidth]{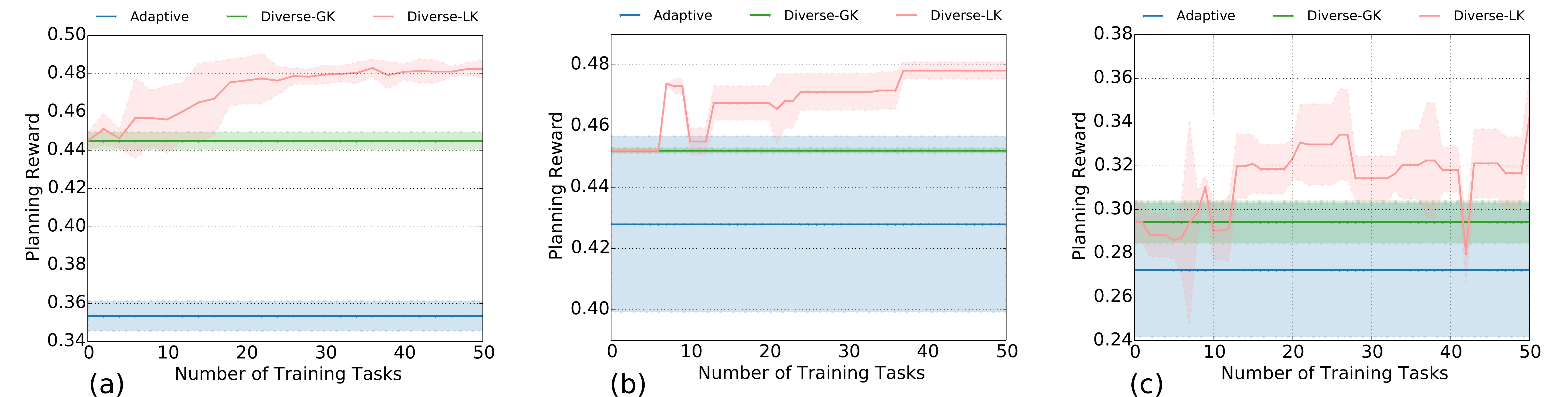}
\caption{The mean learning curve of reward $J(\phi)$ (with 1.96
  standard deviation) as a function of the number of training tasks in
  three domains: 
  (a) Task I: pushing an object off the
  table (b) Task II: pouring into a cup next to a wall (c) Task III: picking up a cup in
  a holder and pour into a cup next to a wall.} 
\label{fig:alltask}
\vskip -0.2in
\end{figure*}
We define the planning reward of a sampler to be
$J_k(\phi) =\sum_{n=1}^\infty s(\phi, n)\gamma^n$, where $s(\phi, n)$
is the indicator variable that the $n$-th sample from $\phi$ helped
the planner to generate the final plan for a particular task instance
$k$. The reward is discounted by $\gamma^n$ with $0< \gamma < 1$, so
that earlier samples get higher rewards (we use $\gamma=0.6$). We
average the rewards on tasks drawn from a predefined distribution,
and effectively report a lower bound on $J(\phi)$, by setting a time
limit on the planner. 
 
The first set of tasks (Task I) we consider is a simple controlled
example where the goal is to push an object off a 2D table with the
presence of an obstacle on either one side of the table or the other
(both possible situations are equally likely).  The presence of these
obstacles is not represented in the context of the sampler, but the
planner will reject sample action instances that generate a collision
with an object in the world and request a new sample. %
We use a fixed range of
feasible actions %
sampled from two rectangles in 2D of unequal sizes. 
The optimal strategy is to first randomly sample from one side of the
table and if no plan is found, sample from the other side.

We show the learning curve of \lk with respect to the planning reward 
metric $J(\phi)$ in Fig.~\ref{fig:alltask} (a). 1000 initial
arrangements of obstacles were drawn randomly for testing. 
 We also repeat the experiments 5 times to obtain the $95\%$ confidence 
 interval. 
For \gk, the kernel inverse is initialized
as $[1,1]$ and if, for example, it sampled on the left side of the
object (pushing to the right) and the obstacle is on the right, it may
not choose to sample on the right side because the kernel indicates
that the other feature is has more diversity. However, after a few
planning instances, \lk is able to figure out the right
configuration of the kernel and its sampling strategy becomes the
optimal one. 

We also tested these three sampling algorithms on two more complicated
tasks. We select a fixed test set with 50 task specifications and
repeat the evaluation 5 times.  The first one (Task II) involves
picking up cup A, getting water from a faucet, move to a pouring
position, pour water into cup B, and finally placing cup A back in its
initial position. Cup B is placed randomly either next to the wall on
the left or right. The second task is a harder version of Task II,
with the additional constraint that cup A has a holder and the sampler
also has to figure out that the grasp location must be close to
the top of the cup (Task~III).

We show the learning results in Fig.~\ref{fig:alltask} (b) and (c)  
and timing results in Tab.~\ref{tb:timing} (after training). 
We conjecture that the sharp turning points in the learning curves of 
Tasks II and III are a result of high penalty on the kernel lengthscales and 
the limited size (50) of the test tasks, and we plan to investigate more 
in the future work.  
Nevertheless, \lk is still able to find a better
solution than the alternatives in Tasks II and III. 
Moreover, the two diverse sampling
methods achieve lower variance on the success rate and perform more
stably  after training.

\hide{
\begin{table}[t]
\caption{Timing results and plan success rates (SR) for pouring task. The runtime only includes runs that successfully completed within 60s.}
\label{tb:timingpour_hard}
\vskip -0.2in
\begin{center}
\begin{footnotesize}
\begin{tabular}{lccc}%
\hline
\abovestrut{0.20in}\belowstrut{0.10in}
Method & Runtime (s) & 60s SR  (\%) & 6s SR (\%) \\
\hline
\abovestrut{0.20in}
Adaptive   & $3.22\pm 6.51$  & $91.0\pm 2.7$& $82.4\pm 5.6$ \\
Diverse-GK & $2.06\pm 1.76$        &  {\color{red} $95.0\pm 1.8 $} & $93.6\pm 2.2 $\\
Diverse-LK  & {\color{red}$1.71\pm 1.23$} & {\color{red} $95.0\pm 1.8$} &  {\color{red}$94.0\pm 1.5$}       \\
\hline
\end{tabular}

\end{footnotesize}
\end{center}
\vskip -0.2in
\end{table}

\begin{table}[t]
\caption{Timing results and plan success rates (SR) for pouring task with a holder around the picked cup. The runtime only includes runs that successfully completed within 60s.}
\label{tb:timingpour}
\vskip -0.2in
\begin{center}
\begin{footnotesize}
\begin{tabular}{lccc}%
\hline
\abovestrut{0.20in}\belowstrut{0.10in}
Method & Runtime (s) & 60s SR  (\%) & 6s SR (\%) \\
\hline
\abovestrut{0.20in}
Adaptive   & $5.79\pm 11.04$  & $51.4\pm 3.3$& $40.9\pm 4.1$ \\
Diverse-GK & $3.90\pm 5.02$        &  $56.3\pm 2.0 $ & $46.3\pm 2.0 $\\
Diverse-LK  & $4.30\pm 6.89$ & {\color{red} $59.1\pm 2.6$} &  {\color{red}$49.1\pm 2.6$}       \\
\hline
\end{tabular}
\end{footnotesize}
\end{center}
\vskip -0.2in
\end{table}
}
\hide{
\begin{figure}
\centering
\includegraphics[width=0.8\columnwidth]{figs/toy_pushing_off_table_maxlen_10_condvar_se_online.pdf}
\caption{The success rate of a planning task with different number of samples generated by a uniform sampler (Uniform), a diverse sampler (Diverse), and an enhanced diverse sampler (Enhanced Diverse). The planning task is to push an object off the table either to the left side or the right side.}
\label{fig:toypush}
\end{figure}
\begin{figure}
\centering
\includegraphics[width=0.8\columnwidth]{figs/haswater_maxlen_10_condvar_se_online.pdf}
\caption{The success rate of a planning task with different number of samples generated by a uniform sampler (Uniform), a diverse sampler (Diverse), and an enhanced diverse sampler (Enhanced Diverse). The goal of the task is to get water from a faucet using an empty cup, pour water into an empty kettle using the cup and place the cup back to its initial position.}
\label{fig:pourtask}
\end{figure}

\begin{figure}
\centering
\includegraphics[width=0.8\columnwidth]{figs/tricky_maxlen_10_condvar_matern_online.pdf}
\caption{The success rate of a planning task with different number of samples generated by a uniform sampler (Uniform), a diverse sampler (Diverse), and an enhanced diverse sampler (Enhanced Diverse). The goal of the planning task is the same as Fig.~\ref{fig:pourtask} except that the cup to be picked has a tall holder and the grasping position has to be above the holder.}
\label{fig:trickypour}
\end{figure}
}

\subsection{Integrated system}
Finally, we integrate the learned action sampling models for pour and 
scoop  with 7 pre-existing robot operations (move, push, pick, place,
fill, dump, stir) in a domain specification for \stripstream.  The robot's
goal is to ``serve'' a cup of coffee with cream and sugar by placing
it on the green coaster near the edge of the table.  Accomplishing
this requires general-purpose planning, including picking where to
grasp the objects, where to place them back down on the table, and what
the pre-operation poses of the cups and spoon should be before
initiating the sensorimotor primitives for pouring and scooping should
be.  Significant perturbations of the object arrangements are handled
without difficulty\footnote{We use the focused algorithm within STRIPStream, 
and it solves the task in 20-40 seconds for a range of different arrangements of objects.}.  
A resulting plan and execution sequence is shown
in the accompanying video at \url{https://youtu.be/QWjLYjN8axg}. 

This work illustrates a critical ability:  to augment the existing
competences of a robotic system (such as picking and placing objects)
with new sensorimotor primitives by learning probabilistic models of
their preconditions and effects and using a state-of-the-art
domain-independent continuous-space planning algorithm to combine them
fluidly and effectively to achieve complex goals.

{
\bibliographystyle{IEEEtran}
\bibliography{IEEEabrv,refs}
}
\end{document}